\documentclass{article}

\usepackage{arxiv}
\usepackage{amsmath}
\usepackage[utf8]{inputenc} 
\usepackage[T1]{fontenc}    
\usepackage{hyperref}       
\usepackage{url}            
\usepackage{booktabs}       
\usepackage{amsfonts}       
\usepackage{nicefrac}       
\usepackage{microtype}      
\usepackage{lipsum}		
\usepackage[all]{nowidow}
\usepackage{graphicx}
\usepackage{float}
\usepackage{placeins}
\usepackage{natbib}
\usepackage{doi}

\title{AurigaNet: A Real-Time Multi-Task Network for Enhanced Urban Driving Perception}

\author{\href{https://orcid.org/0009-0000-6304-7170}{\includegraphics[scale=0.06]{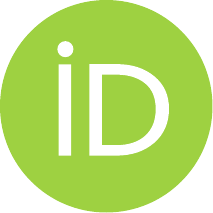}\hspace{1mm}Kiarash Ghasemzadeh}\\
University of Alberta\\
\texttt{kghasemz@ualberta.ca}\\
Shahid Beheshti University\\
\texttt{k.ghasemzadeh@mail.sbu.ac.ir}\\
\And
\href{https://orcid.org/0000-0003-1337-6050}{\includegraphics[scale=0.06]{orcid.pdf}\hspace{1mm}Sedigheh Dehghani}\\
Shahid Beheshti University\\
\texttt{s\_dehghani@sbu.ac.ir}\\
}

\date{}

\renewcommand{\headeright}{}
\renewcommand{\undertitle}{}
\renewcommand{\shorttitle}{AurigaNet}

\hypersetup{
pdftitle={A template for the arxiv style},
pdfsubject={q-bio.NC, q-bio.QM},
pdfauthor={David S.~Hippocampus, Elias D.~Striatum},
pdfkeywords={First keyword, Second keyword, More},
}

\begin{document}
\maketitle

\begin{abstract}
Self-driving cars hold significant potential to reduce traffic accidents, alleviate congestion, and enhance
urban mobility. However, developing reliable AI systems for autonomous vehicles remains a substantial
challenge. Over the past decade, multi-task learning has emerged as a powerful approach to address
complex problems in driving perception. Multi-task networks offer several advantages, including
increased computational efficiency, real-time processing capabilities, optimized resource utilization, and
improved generalization. In this study, we present \textbf{AurigaNet}, an advanced multi-task network
architecture designed to push the boundaries of autonomous driving perception. AurigaNet integrates
three critical tasks: object detection, lane detection, and drivable area instance segmentation. The
system is trained and evaluated using the BDD100K dataset, renowned for its diversity in driving
conditions. Key innovations of AurigaNet include its end-to-end instance segmentation capability, which
significantly enhances both accuracy and efficiency in path estimation for autonomous vehicles.
Experimental results demonstrate that AurigaNet achieves an 85.2\% IoU in drivable area segmentation,
outperforming its closest competitor by 0.7\%. In lane detection, AurigaNet achieves a remarkable 60.8\%
IoU, surpassing other models by more than 30\%. Furthermore, the network achieves an mAP@0.5:0.95
of 47.6\% in traffic object detection, exceeding the next leading model by 2.9\%. Additionally, we validate
the practical feasibility of AurigaNet by deploying it on embedded devices such as the Jetson Orin NX,
where it demonstrates competitive real-time performance. These results underscore AurigaNet's
potential as a robust and efficient solution for autonomous driving perception systems. The code can be found here \url{https://github.com/KiaRational/AurigaNet}.
\end{abstract}

\keywords{Autonomous Vehicles \and Multi-Task Learning \and Path Detection \and Object Detection \and Lane Detection \and Instance segmentation \and Discriminative Features}

\section{Introduction}
\label{intro}

Autonomous vehicles could play an important role in reducing accidents on the road by eliminating human error, which is almost always at fault in any kind of traffic incident. Inherently, autonomous vehicles are designed to avoid very common mistakes made by human drivers, significantly bringing down the occurrence of accidents. Moreover, autonomous vehicles have the potential to significantly improve road safety by enhancing traffic management and reducing injuries and fatalities. To do all of this with associated benefits, self-driving cars should be within affordable costs and show high reliability, hence accessible and trustworthy. This is particularly significant, as human lives depend on their accuracy and reliability \cite{Bagloee2016}.

In recent years, autonomous driving research has focused on developing robust panoptic driving perception systems. These systems integrate tasks like object detection, drivable area segmentation, and lane line segmentation by using onboard sensors to gather essential information about a vehicle's surroundings. This information is then used for higher-level decision-making and behavioral control in self-driving vehicles to ensure safe and reliable operation.

At present, these systems rely on onboard sensors such as cameras and LiDAR. While LiDAR is insensitive to color and light, it is expensive. On the other hand, cameras capture rich texture and color information suitable for object detection and segmentation tasks, and they are cost-effective and easy to install. Therefore, using images captured by cameras along with deep learning models in developing a panoptic driving perception system provides a competitive and cost-effective solution for low-cost ADAS with high efficiency \cite{Yurtsever2020}. 

Neural network-based target detection and image segmentation models have expanded the possibilities
beyond traditional image processing methods. Recent advancements in object detection methods can
be broadly categorized into two approaches: two-stage and single-stage detectors. Two-stage
approaches, such as the RCNN series \cite{maskrcnn,fastrcnn,fasterrcnn,hybridfasterrcnn}, prioritize detection accuracy through a two-step process, albeit at the cost of computational efficiency. On the other hand, single-stage detectors, like the YOLO series \cite{yolo,yolov3,yolov7} and SSD \cite{SSD}, are favored in the industry for their efficiency on embedded devices, balancing speed and accuracy. Hence, multi-task detection networks based on a single model are considered efficient solutions for low-cost ADAS applications.

The segmentation models have also evolved, from UNet \cite{unet} to PSPNet \cite{pspnet}, both of which significantly improved the segmentation tasks. However, while these approaches have achieved success individually, employing separate models for different tasks in practical autonomous driving systems, particularly low-cost ADAS, is often impractical. Implementing distinct models for each task demands substantial computational resources, complicates network deployment, and
contradicts the low-cost imperative. As a result, multi-task networks capable of handling multiple tasks
simultaneously have garnered significant attention in the field of ADAS.

\begin{figure}[H]
    \centering
    \includegraphics[width=0.35\textwidth]{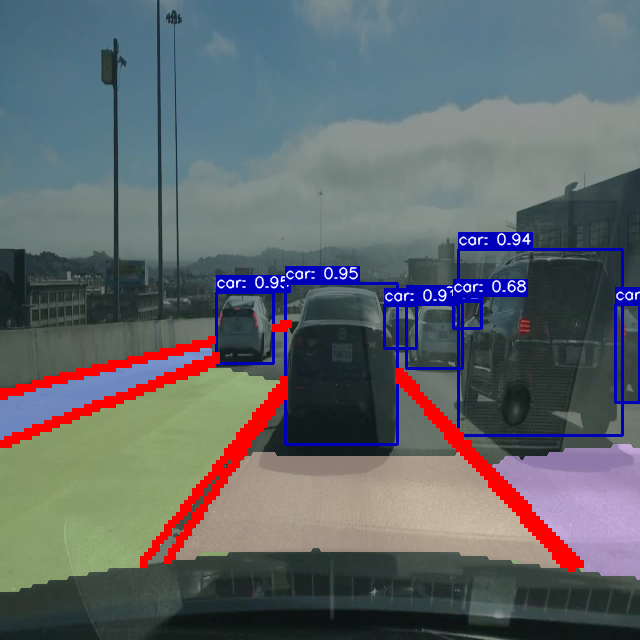}
    \caption{The inference result of AurigaNet}
    \label{fig:1}
\end{figure}

Path estimation fundamentally relies on drivable area instance segmentation, since differentiating between multiple drivable areas or parallel paths is crucial for accurate trajectory planning. Pure semantic segmentation methods cannot provide this distinction, as they assign the same class label to all drivable regions without separating them into distinct instances. Although models like UNet, PSPNet, YOLOP \cite{yolop}, and HybridNets \cite{hybridnets} can detect drivable areas, none of them support end-to-end instance segmentation.
Instead, these models require additional post-processing clustering methods to integrate instance
segmentation. A significant challenge with clustering methods is that the number of clusters in driving
scenes is often unknown, rendering traditional algorithms like KMeans \cite{kmeans} and KNN \cite{knn} clustering methods cannot be applied. While some clustering algorithms, like DBSCAN \cite{dbscan}, can cluster drivable areas without requiring a predefined number of clusters, they are computationally slow.
This limitation directly motivates the design of AurigaNet. To achieve robust instance-level separation of drivable areas without costly clustering, AurigaNet combines two key components: (i) a discriminative loss \cite{discriminative}, which encourages feature embeddings of the same drivable instance to cluster tightly while pushing apart embeddings from different instances, and (ii) deformable convolutions \cite{deformable}, which adapt receptive fields dynamically to align with irregular lane geometries and free-space boundaries. Together, these design choices enable AurigaNet to perform end-to-end drivable area instance segmentation, delivering the reliability required for path estimation while maintaining real-time performance on embedded platforms.
In this study, we present \textbf{AurigaNet}, a cutting-edge multi-task network architecture trained on the challenging BDD100K dataset \cite{bdd100k}. AurigaNet achieves state-of-the-art performance in object detection, lane detection, and drivable area instance segmentation. The inference results of AurigaNet are depicted in Figure \ref{fig:1}.

\newpage
The main contributions of this paper are as follows:

\begin{enumerate}
    \item  We propose AurigaNet, a multi-task deep neural network capable of performing end-to-end drivable area instance segmentation, object detection, and lane detection, all within a single unified architecture. Unlike existing networks such as YOLOP \cite{yolop} and HybridNets \cite{hybridnets},
AurigaNet eliminates the need for additional post-processing steps for path estimation, making it a more efficient
solution for path estimation and planning in ADAS systems.
    \item We enhance AurigaNet by integrating discriminative loss function \cite{discriminative} and deformable convolution \cite{deformable} for better instance separation in drivable area segmentation, improving accuracy in complex environments. These innovations enhance AurigaNet’s accuracy and flexibility across diverse driving scenarios.
    \item We demonstrate that AurigaNet achieves state-of-the-art performance in drivable area segmentation (85.2\% IoU), lane detection (60.8\% IoU), and traffic object detection (47.6\% mAP).
    \item Extensive experiments validate the real-time performance of AurigaNet on embedded devices, such as the Jetson Orin NX, proving its practical feasibility for low-cost ADAS applications.
\end{enumerate}
The paper’s organization is as follows: Section \ref{rw} introduces the relevant work on drivable area detection, lane detection, and traffic object detection. Section \ref{met} provides a detailed description of the proposed model, AurigaNet. The
dataset, experimental setup, and experimental findings are all described in Section \ref{test}. Finally, the conclusion is presented in Section \ref{conc}.

\section{Related Work}
\label{rw}

Recently, the field of deep learning has experienced significant advancements in driving perception, especially in traffic object detection, lane detection, drivable-area segmentation, and multi-task learning. In this section, we examine several representative works to highlight the progress in autonomous driving perception.

\subsection{Traffic Object Detection}

Object detection has evolved from traditional region-proposal methods to deep learning-driven approaches. The introduction of Convolutional Neural Networks led to region-based CNNs like R-CNN \cite{maskrcnn}, Fast R-CNN \cite{fastrcnn}, and Faster R-CNN \cite{fasterrcnn}, which established high-accuracy baselines. One-stage detectors, such as SSD \cite{SSD}, YOLO \cite{yolo}, YOLOv3 \cite{yolov3}, YOLOv4 \cite{yolov4}, and YOLOv7 \cite{yolov7}, improved efficiency by directly predicting object bounding boxes and class probabilities. Innovations like anchor mechanisms and Feature Pyramid Networks (FPN \cite{lin2017feature}) enhanced detection accuracy across multiple scales. AurigaNet adopts a one-stage detection head similar to YOLO, which is essential for real-time applications.

\subsection{Lane Detection and Drivable-Area Segmentation}

Lane detection and drivable-area segmentation provide complementary structural cues for trajectory planning. LaneNet \cite{wang2018lanenet} reformulates lane detection as an instance segmentation problem, enabling flexible handling of variable lane counts and complex road geometries. PINet \cite{ko2021key} uses a point-based representation to simultaneously estimate key point coordinates and perform instance segmentation, reducing false positives. Lightweight designs and knowledge distillation methods have also been proposed to meet real-time constraints \cite{sad}. 

Drivable-area segmentation is typically formulated as semantic segmentation, offering pixel-level precision. Early models like FCN \cite{long2015fully} laid the foundation but struggled with low-resolution outputs. Architectures such as PSPNet \cite{pspnet} and UNet \cite{unet} improved accuracy, though they can be computationally demanding. Traditional drivable path estimation often relies on clustering-based post-processing, such as k-means \cite{kmeans}, kNN \cite{knn}, DBSCAN \cite{dbscan}, Mean-Shift \cite{meanshift}, or Von-Mises shift \cite{von_mis}, which introduces latency and potential instability. AurigaNet addresses this by performing drivable-area instance segmentation end-to-end.


\subsection{Multi-Task Networks}

Multi-task learning unifies multiple perception outputs within a single architecture, reducing redundancy and runtime \cite{caruana1997multitask}. Early multi-task autonomous driving systems, such as YOLOP \cite{yolop} and HybridNets \cite{hybridnets}, demonstrated the feasibility of combining object detection, lane detection, and drivable-area segmentation within one model. 
Transformer-based approaches, such as BEVFormer \cite{bevformer}, learn spatiotemporal bird’s-eye-view representations for multi-view 3D detection and map segmentation. UniAD \cite{uniad} further integrates perception, prediction, and planning in a single query-based framework, supporting planning-aware training and inference. These methods achieve strong performance but often come with high computational costs, motivating lighter-weight solutions for embedded ADAS, which AurigaNet addresses.

\begin{figure}[H]
\includegraphics[width=1\textwidth]{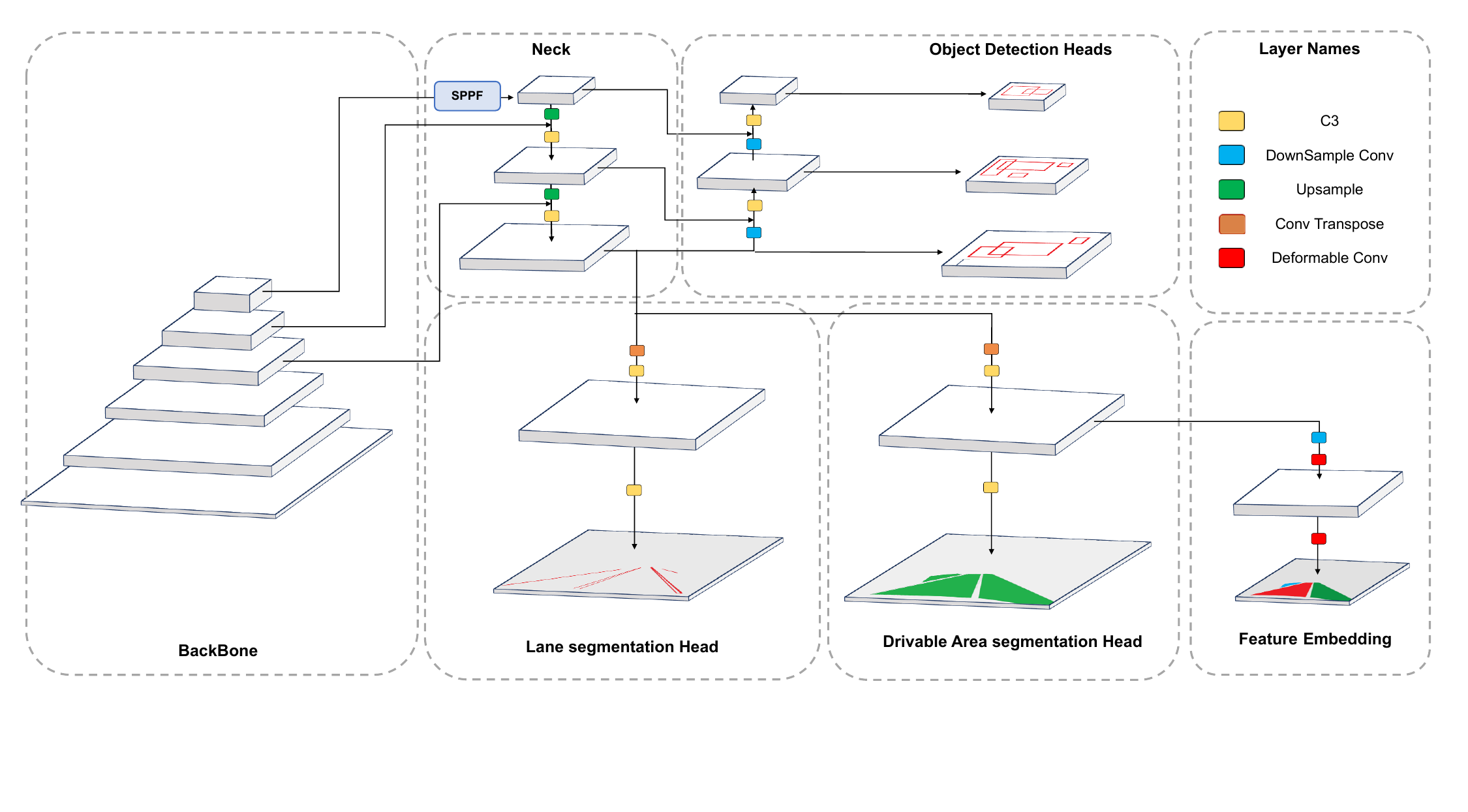}
\caption{The architecture of AurigaNet}
\label{fig:2}       
\end{figure}

\section{Method}
\label{met}

This study introduces AurigaNet, an end-to-end feed-forward network architecture inspired by YOLOP\cite{yolop} and HybridNets\cite{hybridnets}. Building upon the core design of these architectures, AurigaNet incorporates a more robust architecture and a feature embedding head for drivable area instance segmentation. As depicted in Figure \ref{fig:2}, the proposed network comprises a single shared encoder, consisting of a backbone and a neck, followed by three dedicated decoder heads for drivable Area Instance Segmentation, traffic object detection, and lane detection. A detailed diagram of AurigaNet's architecture is provided in the Appendix section.

\subsection{Shared Encoder}
The shared encoder extracts comprehensive features from the input image. This two-stage process involves a backbone network for feature extraction and a neck network for feature manipulation.

\subsubsection{Backbone}
The backbone network, tasked with extracting informative features from the input image, plays a pivotal role in reducing the image's spatial resolution while simultaneously increasing its depth (number of channels). Similar to YOLOP \cite{yolop}, AurigaNet employs the CSPDarknet architecture, known for addressing the issue of gradient duplication during optimization. This choice enables efficient feature propagation and reuse, leading to a reduction in both network parameters and computational requirements. Consequently, it contributes significantly to achieving real-time performance, a crucial aspect for practical autonomous driving applications.

\subsubsection{Neck}
The neck network collects and fuses features extracted from various stages of the backbone network. This process, commonly referred to as feature fusion, is achieved through concatenation, akin to the YOLO architecture. Notably, the neck network in AurigaNet leverages two key modules:\\
\begin{itemize}

\item Spatial Pyramid Pooling Fusion (SPPF): This module integrates features across different scales, enabling the network to capture information at diverse levels of granularity, resulting in a more comprehensive understanding of the scene.

\item Feature Pyramid Network (FPN): This module helps with the integration of features from various semantic levels within the network. This integration process enhances the network's ability to localize and classify objects at different scales.
\end{itemize}

\subsection{Decoders}
Each task has a dedicated decoder, carefully designed to accurately extract information relevant to that task.

\subsubsection{Traffic Object Detection Head}
In alignment with the YOLOv5 model \cite{yolov5}, our approach adopts an anchor-based multi-scale detection scheme. This involves utilizing the Path Aggregation Network (PAN), a bottom-up feature pyramid network, in conjunction with the top-down transfer of semantic features facilitated by the Feature Pyramid Network (FPN). The fusion of these pathways optimizes feature integration, with PAN's multi-scale feature maps directly used for detection. Each grid in the map gets three anchors with different aspect ratios. The detection head predicts position offsets, size scaling, category probabilities, and confidence for each prediction, ensuring robust detection.
\vspace{-10pt}
\subsubsection{Drivable Area Instance Segmentation}
\begin{figure}[H]
    \centering
    \includegraphics[width=0.6\linewidth]{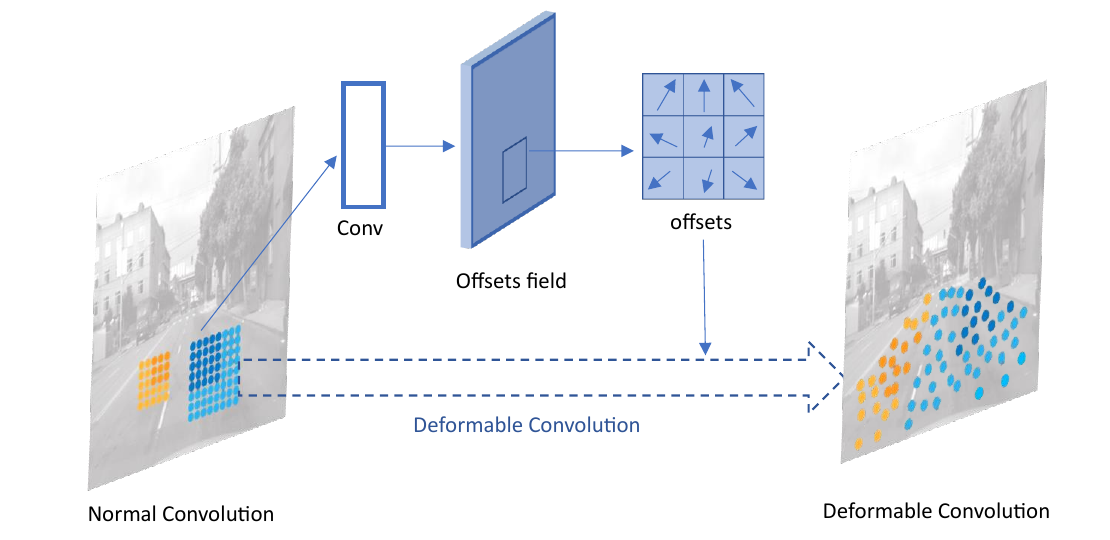}
    \caption{Illustration of offset field role in Deformable Convolution that helps to align with the shape of instances}
    \label{fig:4}
\end{figure}

Drivable-area detection involves two decoders: a binary segmentation head and a feature-embedding head. Now, we generate the binary segmentation head by using the bottom layer of the FPN with dimensions ($W/8\times H/8\times C$). Then, the segmentation branch undergoes one transpose convolution and multiple C3 layers (defined in Figure \ref{fig:12}) to recover the output feature map into ($W/4\times H/4\times 1$), which would capture the probability distribution for every pixel in the input image and separate the drivable area from the background.
For the Feature Embedding layer, we take the first upsampled feature map of the binary segmentation head, which is ($W/4\times H/4\times C$), and apply a down-sample layer so that the network can learn more features from the last layer. Then, we make use of the deformable convolution layers \cite{deformable}, which add 2D offsets to the regular grid sampling locations in standard convolution and enable free-form deformation of the sampling grid. The offsets are learned from the previous feature maps via additional convolutional layers, which are specifically added to make the deformation local, dense, and adaptive, as shown in Figure \ref{fig:4}. This helps distinguish different instances and outputs a feature embedding map for instance segmentation. The size of the output will be ($W/8\times H/8\times 8$).

\subsubsection{Lane Detection}
\begin{figure}[H]
    \centering
    \includegraphics[width=\textwidth]{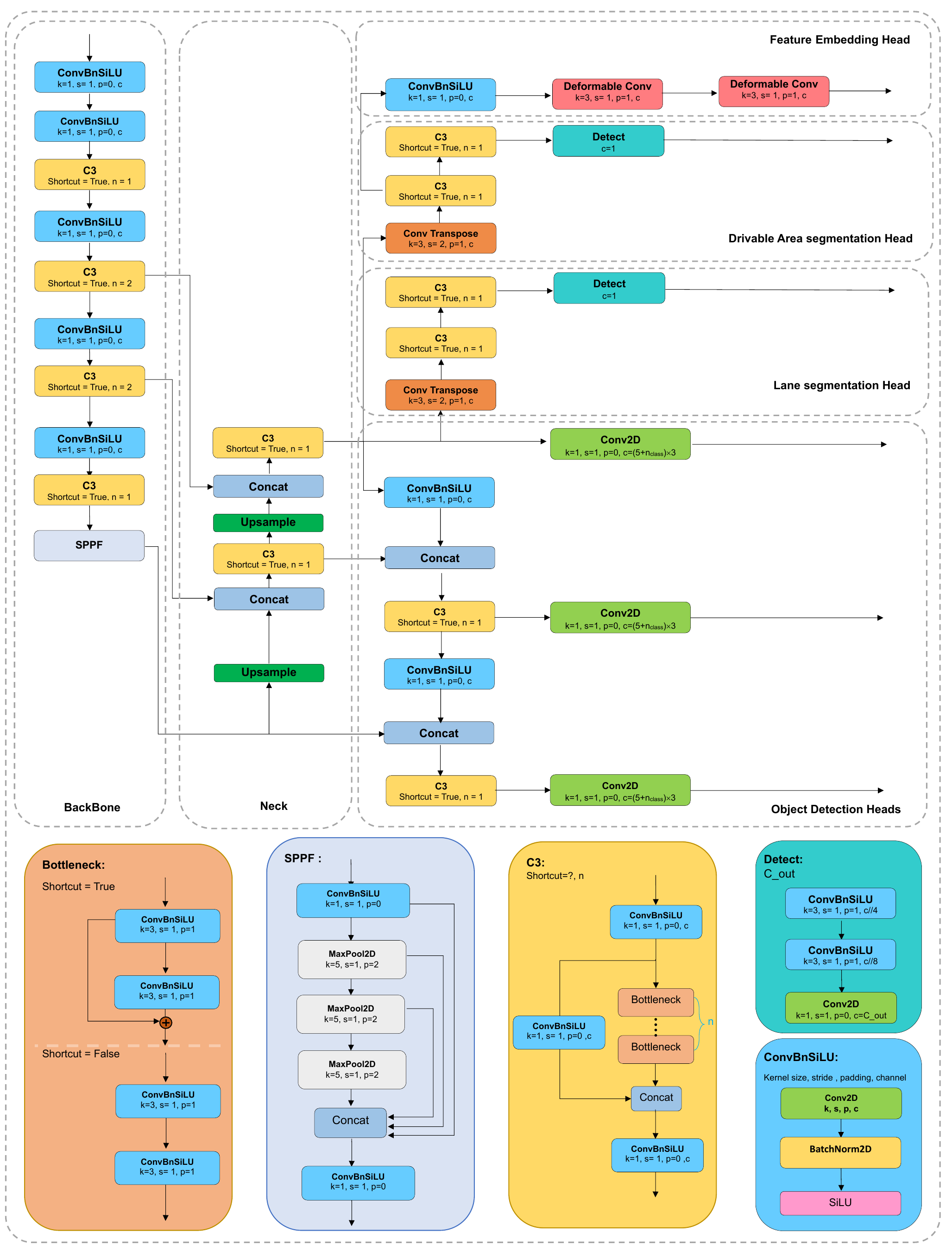}
    \caption{The detailed architecture of the AurigaNet model.}
    \label{fig:12}
\end{figure}

The process of lane detection involves a binary segmentation head. Same as the drivable area head to create the binary segmentation head, we use the bottom layer of the Feature Pyramid Network (FPN) with dimensions ($W/8\times H/8\times C$). The segmentation branch then undergoes one transpose convolution and C3 layers (defined in Figure \ref{fig:12}) to restore the output feature map to dimensions of ($W/4\times H/4\times 1$).

\subsection{Loss functions}
The total loss function in our system is computed as per the equation \ref{eq:1}, where $\gamma_1$, $\gamma_2$, and $\gamma_3$ are the weightage factors assigned to object detection, drivable area, and lane respectively. Each head in our system has a different loss function. \\
\begin{equation} \label{eq:1}
    L_{Total} = \gamma_{1}L_{objdetection}+\gamma_{2}L_{area}+\gamma_{3}L_{lane}
\end{equation}
\begin{equation} \label{eq:2}
    L_{objdetection} = \alpha_{1}L_{box}+\alpha_{2}L_{obj}+\alpha_{3}L_{class}
\end{equation} \\
For object detection, we employ the loss function defined in equation \ref{eq:2}, similar to YOLOv5 \cite{}. The loss function consists of three components: $L_{box}$, $L_{obj}$, and $L_{class}$, each weighted by $\alpha_1$, $\alpha_2$, and $\alpha_3$ respectively. The object and class losses are computed using binary cross-entropy, while $L_{box}$ is calculated using $L_{SIOU}$ \cite{siou}, which incorporates multiple geometric considerations to improve performance.
To calculate the loss for drivable area segmentation, we use the Dice loss and binary cross-entropy (BCE) loss, which are combined with $\alpha_{21}$ and $\alpha_{22}$ respectively. Additionally, we utilize a feature embedding head to extract features from the predicted drivable area mask. The discriminative loss function \cite{discriminative} as defined in equations \ref{eq:3}, \ref{eq:4}, \ref{eq:5}, and \ref{eq:6}, is designed to enhance the separation between different instances while maintaining compactness within each instance. The process achieves its goals by optimizing the feature space in two key ways. First, it minimizes the intra-class variance, which brings features of the same instance closer together. This grouping improves the clarity and definition of each instance. Second, the discriminative loss maximizes the inter-class variance, pushing features of different instances further apart. This clear separation is crucial for accurately differentiating between multiple instances that may be closely positioned, a common challenge in autonomous driving scenarios. For more details on the specific tuning of these parameters and their effects, please refer to the original paper \cite{discriminative}. Ultimately, all the losses are combined with appropriate alpha weights to calculate the total loss for drivable area segmentation.

\begin{figure}[H]
    \centering
    \includegraphics[width=0.5\linewidth]{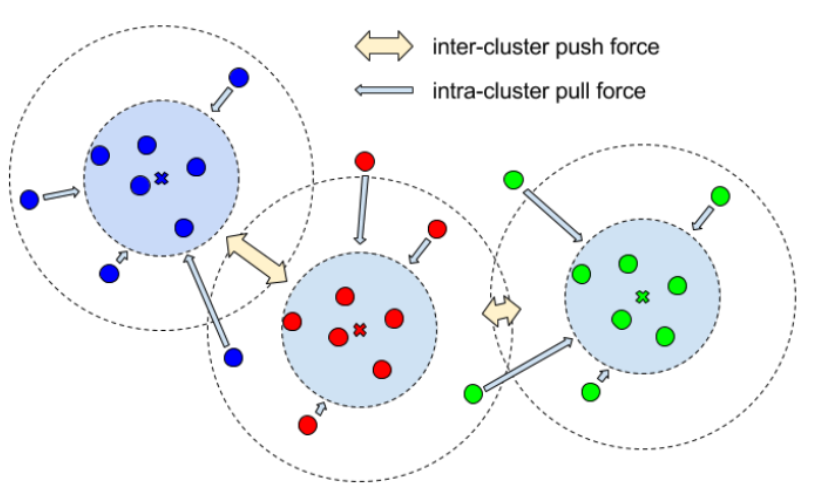}
\caption{Illustration of discriminative loss: same-instance features are clustered, while different instances are separated in feature space.}    \label{fig:52}
\end{figure}

Figure \ref{fig:52} illustrates the intuition behind the discriminative loss: embeddings of pixels belonging to the same instance are pulled together into compact clusters, while embeddings of different instances are pushed apart, ensuring clear separation in feature space.
\begin{equation}\label{eq:3}
L_{discriminative}=\alpha \times L_{v a r}+\beta \times L_{\text {dist }}+\gamma \times L_{r e g}
\end{equation}
\begin{equation}\label{eq:4}
L_{v a r}=\frac{1}{C} \sum_{c=1}^C \frac{1}{N_c} \sum_{i=1}^{N_c}\left[\left\|\mu_c-x_i\right\|-\delta_{\mathrm{v}}\right]_{+}^2
\end{equation}
\begin{samepage}
\begin{equation}\label{eq:5}
L_{dist}=\frac{1}{C(C-1)} \sum_{c_A=1}^C 
\sum_{c_B=1}^C\left[2 \delta_{\mathrm{d}}-\left\|\mu_{c_A}-\mu_{c_B}\right\|\right]_{+}^2 
\end{equation}
\hspace{7.5cm}$_{c_A \neq c_B}$
\begin{equation}\label{eq:6}
L_{r e g}=\frac{1}{C} \sum_{c=1}^C\left\|\mu_c\right\|
\end{equation}
\end{samepage}
\begin{equation} \label{eq:7}
    L_{drivable} = \alpha_{21} L_{dice} + \alpha_{22} L_{BCE} + \alpha_{23} L_{discriminative}
\end{equation} 

Moving on to lane line segmentation, we also use the Dice loss and BCE loss, which are combined with $\alpha_{31}$ and $\alpha_{32}$ respectively. The predicted lane masks are then post-processed to obtain a smooth curve representation of the lanes. Similar to drivable area segmentation, all losses are added together with appropriate alpha weights to obtain the total loss for lane line segmentation.
\begin{equation} \label{eq:8}
    {L_{lane} = \alpha_{31} L_{dice} + \alpha_{32} L_{BCE}}
\end{equation}
\subsection{Synergistic Enhancement for Instance Segmentation}
\begin{figure}[H]
    \centering
    \includegraphics[width=0.6\linewidth]{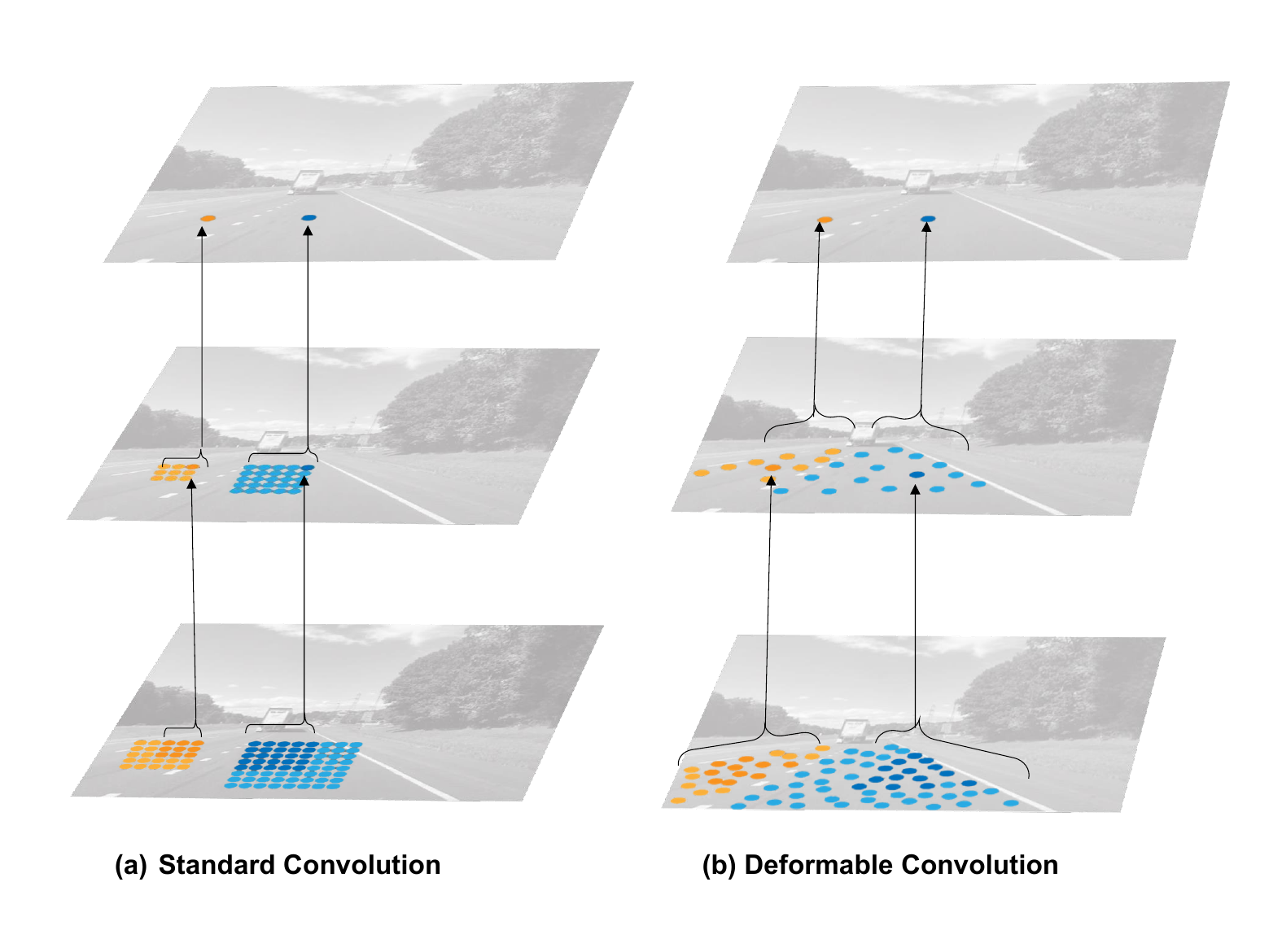}
    \caption{The fixed receptive field in standard convolution and the adaptive receptive field in deformable convolution are illustrated using two layers.}
    \label{fig:5}
\end{figure}

\begin{figure}[H]
    \centering
    \includegraphics[width=1\textwidth]{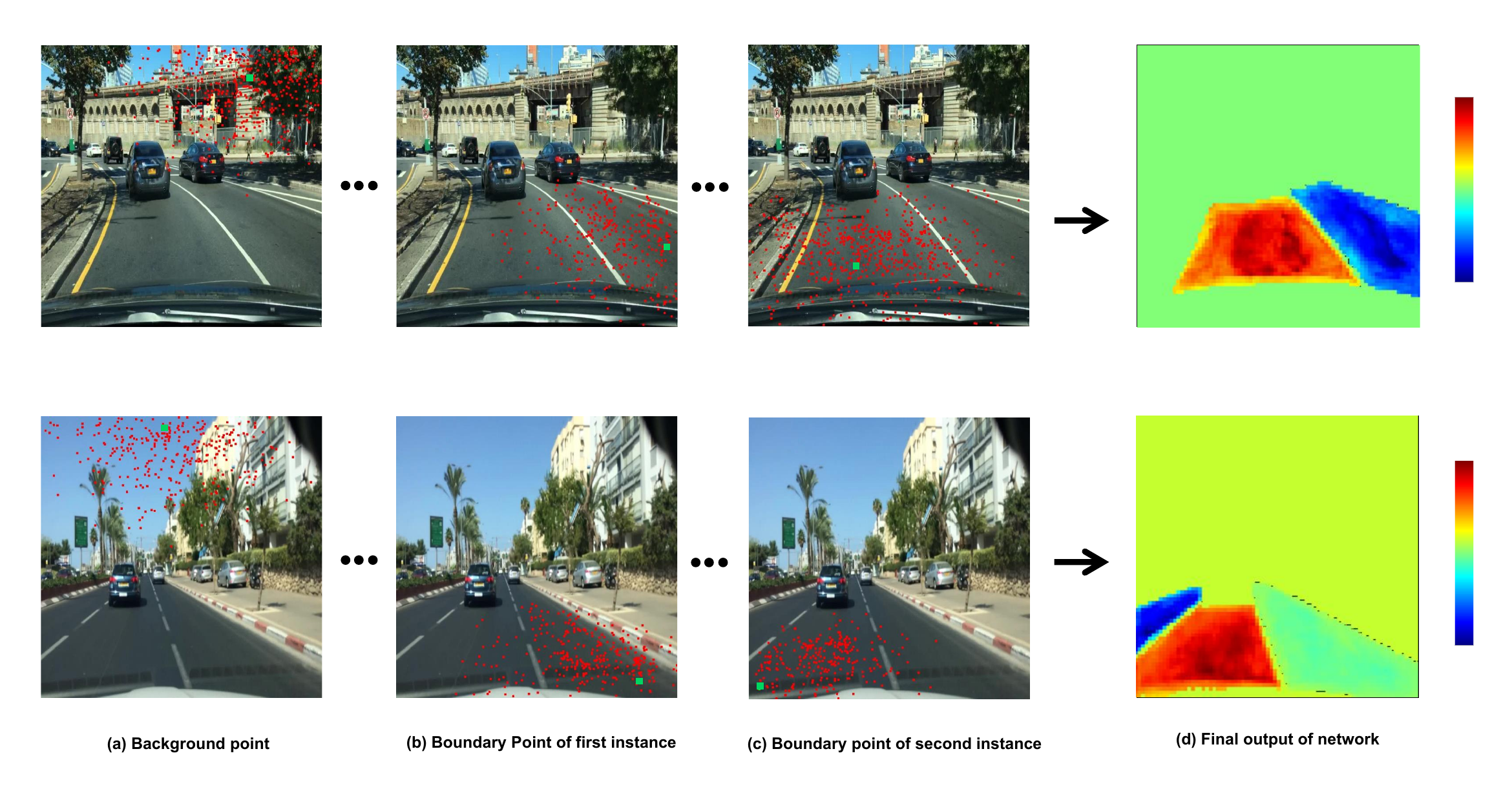}
    \caption{Each set of three images displays the sampling locations ($9^3$ = 729 red points in each image) in three levels of 3 × 3 deformable filters (see Figure \ref{fig:5} as a reference) for three activation units (green points) on the background (a), a first instance (b), and second instance with different shape (c), respectively. The total result of all deformable filters learned by the discriminative loss function is presented in (d).}
    \label{fig:6}
\end{figure}

When the discriminative loss function and stacked deformable convolutions are combined, they create a powerful synergy that significantly improves feature embedding for instance segmentation. The discriminative loss function guides the learning process so that the features of each instance are tightly grouped. Stacked deformable convolutions use this organized feature space to learn offset maps that align the convolutional filters to the center of each instance the concept is illustrated in Figure \ref{fig:5}. By examining the receptive field of each pixel in the output, we observe that this alignment holds true and also performs well at the boundaries of each instance. Examples are depicted in Figure \ref{fig:6}. The offset fields effectively point towards the center of each instance, even at the boundary points, ensuring accurate delineation and improved segmentation quality. This precise alignment enables the network to focus on the exact regions of interest, enhancing segmentation accuracy by capturing the detailed boundaries of each object.

\subsection{Mean Shift Clustering for Embedding Space}

In this work, we employ mean shift clustering \cite{meanshift} for segmentation tasks within our embedding space. Given that our embedding space is normalized, we utilize the von Mises-Fisher (vMF) mean shift algorithm \cite{von_mis}, which is particularly suited for clustering data on a unit hypersphere.

Our feature embedding output has dimensions \( N \times C \times H \times W \), where \( N \) is the batch size, \( C \) is the number of channels, \( H \) is the height, and \( W \) is the width. To apply mean shift clustering, we reshape the feature map \( F \in \mathbb{R}^{H \times W \times C} \) into a feature embedding matrix \( X \in \mathbb{R}^{n \times C} \), where \( n = H \times W \). This matrix \( X \) is then used as input for the clustering algorithm\cite{nvidiargbd}.

The mean shift algorithm iteratively shifts data points towards the mode of the data distribution, effectively identifying clusters without requiring a predefined number of clusters. When using the von Mises-Fisher distribution, the mean shift update for a point \( x_i \) in the embedding space is given by:
\begin{equation} \label{eq:10}
x_i^{(t+1)} = \frac{\sum_{j=1}^{n} x_j \exp(\kappa x_j^\top x_i^{(t)})}{\left\|\sum_{j=1}^{n} x_j \exp(\kappa x_j^\top x_i^{(t)})\right\|}
\end{equation}
where \( \kappa \) is the concentration parameter of the vMF distribution, \( x_j^\top x_i^{(t)} \) denotes the dot product and measures similarity between the current point and other points, and \( t \) represents the iteration step.

The vMF mean shift algorithm aligns data points towards their mean direction on the hypersphere for more accurate clustering in the embedding space. Applying this technique significantly improves instance segmentation accuracy and maintains a reasonable inference time, enhancing the overall performance of our multi-task network.

\FloatBarrier

\section{Experiments}
\label{test}
\begin{figure}[H]
    \centering
    \includegraphics[width=1\linewidth]{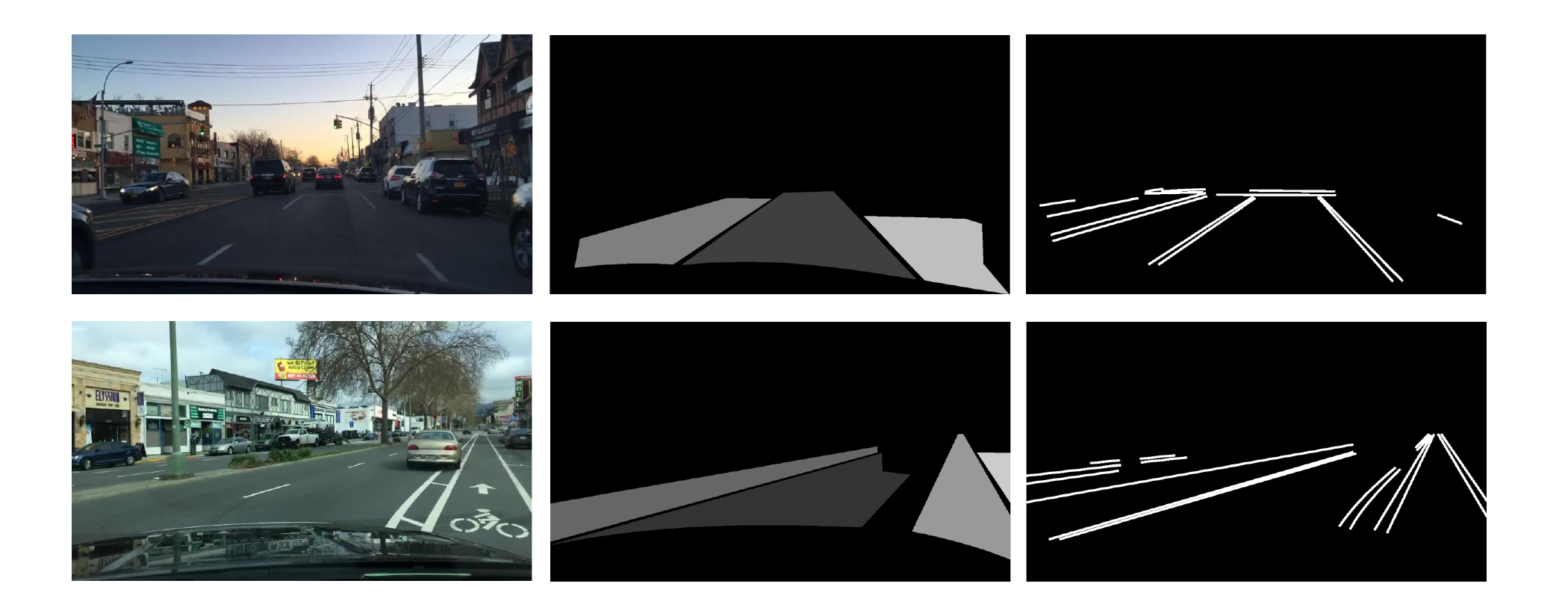}
    \caption{The partial data and annotation of BDD100K dataset}
    \label{fig:7}
\end{figure}
\subsection{BDD100K Dataset and Preprocessing}

The BDD100K dataset \cite{bdd100k} is currently the largest and most comprehensive public driving dataset, consisting of 100,000 images collected from diverse geographic locations and driving conditions across the United States. It is unique in that it provides unified annotations for multiple perception tasks that are essential for autonomous driving, including object detection, drivable area segmentation, and lane detection. 
A major strength of BDD100K lies in its diversity and coverage. The dataset includes scenes recorded under varied lighting conditions (daytime, nighttime, dawn/dusk), weather scenarios (clear, rainy, snowy, foggy), and traffic densities, as well as from different cities and road types (urban centers, highways, residential areas). This diversity provides strong support for model generalizability, as networks trained on BDD100K are naturally exposed to a broad distribution of driving environments.

The dataset is divided into three parts: 70,000 training images, 10,000 validation images, and 20,000 test images. However, since the test set annotations are not yet fully public, we evaluate our network’s performance on the validation set. Representative examples of the data and annotations are shown in Figure \ref{fig:7}.

For preprocessing, images were resized to a standard resolution and normalized to ensure consistency and efficient training. Data augmentation techniques, such as random rotations, motion blur, noise, shadows, flipping, and brightness adjustments, were applied to improve the model's robustness across various driving scenarios.

\subsection{Training Paradigm}
The model was trained using a high-performance computing setup to ensure efficiency. The details of the hardware and software environment are provided in Table \ref{tab:1}.
The hyperparameters listed in table \ref{tab:2} were used for training AurigaNet.
The training process involved an initial warmup phase where the learning rate and momentum were gradually adjusted to stabilize training. This was followed by the main training phase, where the learning rate was dynamically adjusted based on the model's performance.
The hardware setup, data preprocessing, and hyperparameters tuning led to effective training of AurigaNet, resulting in high accuracy and robust performance in driving perception tasks.
\begin{table}[H]
\caption{Training Setups}
\label{tab:1}       
\begin{tabular}{ll}
\hline\noalign{\smallskip}
\textbf{Component} & \textbf{Specification} \\
\noalign{\smallskip}\hline\noalign{\smallskip}
CPU & intel core i5-13600K \\ 
GPU & NVIDIA RTX 4080 \\ 
RAM & 64 GB \\ 
Storage & 2 TB SSD \\
Operating System & Ubuntu 20.04 \\ 
Deep Learning Framework & PyTorch 2.3.1 CUDA 12.1 \\
\noalign{\smallskip}\hline
\end{tabular}
\end{table}
\begin{table}[H]
\caption{Hyperparameters}
\label{tab:2}       
\begin{tabular}{ll}
\hline\noalign{\smallskip}
\textbf{Parameter Name} & \textbf{Value} \\
\noalign{\smallskip}\hline\noalign{\smallskip}
Optimizer & Adam \\ 
Learning Rate & $10^{-4}$ \\ 
Total Epochs & 250 \\
Warmup Epochs & 3 \\ 
Warmup Bias Learning Rate & $10^{-2}$ \\
Warmup Momentum & 0.8 \\ 
Momentum  & 0.937 \\
Batch Size & 16 \\
\noalign{\smallskip}\hline
\end{tabular}
\end{table}

\subsection{Model Evaluation}

For evaluating the performance of AurigaNet on different tasks, we employed several metrics, each tailored to the specific task at hand.

\subsubsection{Drivable Area Instance Segmentation and Lane Segmentation}

For drivable area and lane segmentation evaluation, we used Intersection over Union (IoU).
For evaluating the drivable area instance segmentation performance we used mAP50.
\begin{equation} \label{eq:miou}
IoU = \frac{TP}{TP + FP + FN}
\end{equation}
where \( TP \), \( FP \), and \( FN \) represent the true positives, false positives, and false negatives.
\begin{equation} \label{eq:accuracy}
Accuracy = \frac{TP + TN}{TP + TN + FP + FN}
\end{equation}
Where \( TP \), \( TN \), \( FP \), and \( FN \) are the true positives, true negatives, false positives, and false negatives, respectively.
\subsubsection{Traffic Object Detection}  
For traffic object detection evaluation, we used the mean Average Precision at IoU thresholds from 0.5 to 0.95 (mAP@0.5:0.95) and recall.
\begin{equation} \label{eq:map}
mAP_{50:95} = \frac{1}{N} \sum_{i=1}^{N} AP_i^{50:95}
\end{equation}
where \( mAP_i^{50:95} \) is the average precision computed over multiple IoU thresholds from 0.5 to 0.95 for the \(i\)-th object category.
\begin{equation} \label{eq:recall}
Recall = \frac{TP}{TP + FN}
\end{equation}

\subsection{Comparative Analysis}

In this section, we compare the performance of AurigaNet with other state-of-the-art models on the BDD100K dataset across various tasks using the mentioned metrics.

\subsubsection{Drivable Area Segmentation}
AurigaNet's performance on drivable area segmentation was evaluated and compared with the baseline models using IoU and accuracy. The comparison results of drivable area segmentation are mentioned in Table \ref{tabletotal}. The drivable area instance segmentation performance using mAP50 is mentioned in Table \ref{tabletotal}.
AurigaNet outperformed the baseline models, achieving higher scores across all metrics, indicating superior segmentation performance.

\subsubsection{Lane Detection}

The performance of AurigaNet on lane detection was compared using IoU and accuracy. The comparison results of lane detection are mentioned in Table \ref{tabletotal}.
AurigaNet demonstrated superior performance in lane detection, with higher IoU and accuracy compared to the baseline models.

\subsubsection{Traffic Object Detection}

The evaluation of traffic object detection performance involved comparing AurigaNet with baseline models using mAP50:95 and recall. The comparison results of traffic object detection are mentioned in Table \ref{tabletotal}. We compare and analyze the results of these vehicle detections, as multi-task networks such as YOLOP only focus on the four vehicle categories (car, bus, truck, and train) in the BDD100K dataset for traffic object detection.
AurigaNet achieved higher mAP50:95 demonstrating its effectiveness in detecting traffic objects more accurately than the baseline models.

\begin{table}[H]
\centering
\caption{Comparison of different models on Drivable Area, Lane, and Traffic Object Detection tasks}
\makebox[\textwidth][c]{
\label{tabletotal}
\begin{tabular}{lccccccccccc}
\noalign{\smallskip}\hline\noalign{\smallskip}
 &  & \multicolumn{3}{c}{\textbf{Drivable Area Detection}} & & \multicolumn{2}{c}{\textbf{Lane Detection}} & & \multicolumn{2}{c}{\textbf{Traffic Object Detection}} \\
\hline\noalign{\smallskip}
\textbf{Model}  & \textbf{Input Size} & \textbf{IoU} & \textbf{Accuracy} & \textbf{mAP50} & \multicolumn{1}{c}{\hspace{0.3cm}} & \textbf{IoU} & \textbf{Accuracy} & \multicolumn{1}{c}{\hspace{0.3cm}} & \textbf{mAP@0.5:0.95} & \textbf{Recall} \\
\noalign{\smallskip}\hline\noalign{\smallskip}
FCN \cite{long2015fully}& 769x769 & 74.8 & 90.9 & - & & - & - & & - & - \\
PSPNet \cite{pspnet}& 769x769 & 83.5 & 94.9 & - & & - & - & & - & - \\
\noalign{\smallskip}
SCNN \cite{scnn} & 288x800 & - & - & - & & 15.84 & 35.79 & & - & - \\
Enet-SAD \cite{sad}& 360x640 & - & - & - & & 16.02 & 36.56 & & - & - \\
\noalign{\smallskip}
Yolov5s \cite{yolov5sresult} & 640x640 & - & - & - & & - & - & & 42.5 & 62.5 \\
YOLOP \cite{yolop} & 640x640 & 84.5 & 97.3 & - & & 26.20 & 70.5 & & 43.1 & 89.2 \\
HybridNets \cite{hybridnets} & 640x640 & 83.4 & 96.3 & - & & 31.60 & 85.4 & & 44.7 & \textbf{92.8} \\
\textbf{AurigaNet(Ours)} & 640x640 & \textbf{85.2} & \textbf{97.7} & \textbf{87.25} & & \textbf{60.80} & \textbf{98.77} & & \textbf{47.6} & 75.9 \\
\noalign{\smallskip}\hline
\end{tabular}}
\end{table}
\subsubsection{Time Comparison}
Table \ref{tab_performance} summarizes the network parameters, the network inference in FPS on 4080 GPU, and the total inference in FPS on Jetson Orin NX for the models under consideration.

All models were evaluated in \textbf{FP32} for consistency. In practice, deployment can leverage NVIDIA TensorRT to quantize models to \textbf{FP16} or \textbf{INT8}, typically yielding 1.5--2$\times$ higher throughput with lower memory usage. In this study, we report FP32 results to provide a reproducible baseline.

Although AurigaNet includes a feature embedding output, it demonstrates a competitive balance between network parameters and inference speed, offering a notable trade-off compared to other models. As mentioned in table \ref{tab_performance} AurigaNet achieved higher FPS in Jetson Orin NX embedded device. Jetson Orin NX specification is provided in table \ref{tab:orin NX}.
\begin{table}[H]
\caption{Network Parameters and Inference Time Results}
\label{tab_performance}
\begin{tabular}{lccc}
\hline\noalign{\smallskip}
\textbf{Model}  & \textbf{Parameters(M)} & \textbf{FPS$^{4080}$} & \textbf{FPS$^{NX}$}\\
\noalign{\smallskip}\hline\noalign{\smallskip}
YOLOP & 7.90 & \textbf{362} & 4.002 \\
HybridNets & 12.83 & 139.98 & 1.986 \\
\textbf{AurigaNet} & 9.09 & 217 & \textbf{5.077} \\
\noalign{\smallskip}\hline
\end{tabular}
\end{table}
\begin{table}[H]
\caption{Nvidia Orin Nx Specification}
\label{tab:orin NX}       
\begin{tabular}{ll}
\hline\noalign{\smallskip}
\textbf{Component} & \textbf{Specification} \\
\noalign{\smallskip}\hline\noalign{\smallskip}
CPU & 6-core Arm Cortex-A78AE \\ 
GPU & NVIDIA Ampere 1024 CUDA cores \\ 
Memory &  8GB LPDDR5 \\ 
Power & 10W - 20W \\
Operating System & Ubuntu 20.04 \\
\noalign{\smallskip}\hline
\end{tabular}
\end{table}
\subsubsection{Results Comparison}
\begin{figure}[H]
\centering
\includegraphics[width=0.85\textwidth]{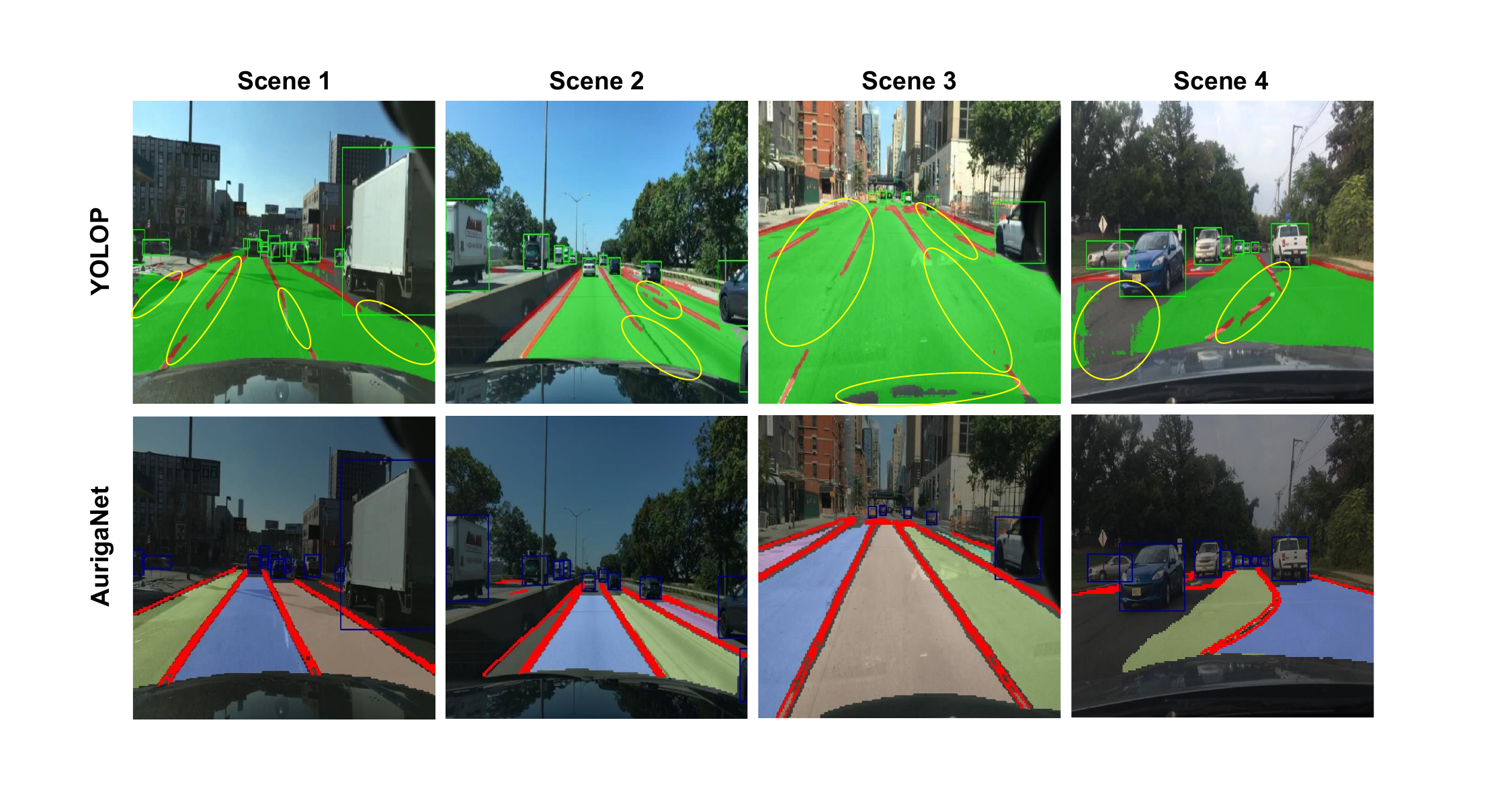}
\caption{The day-time results.}
\label{fig:8}
\end{figure}

\begin{figure}[H]
\centering
\includegraphics[width=0.85\textwidth]{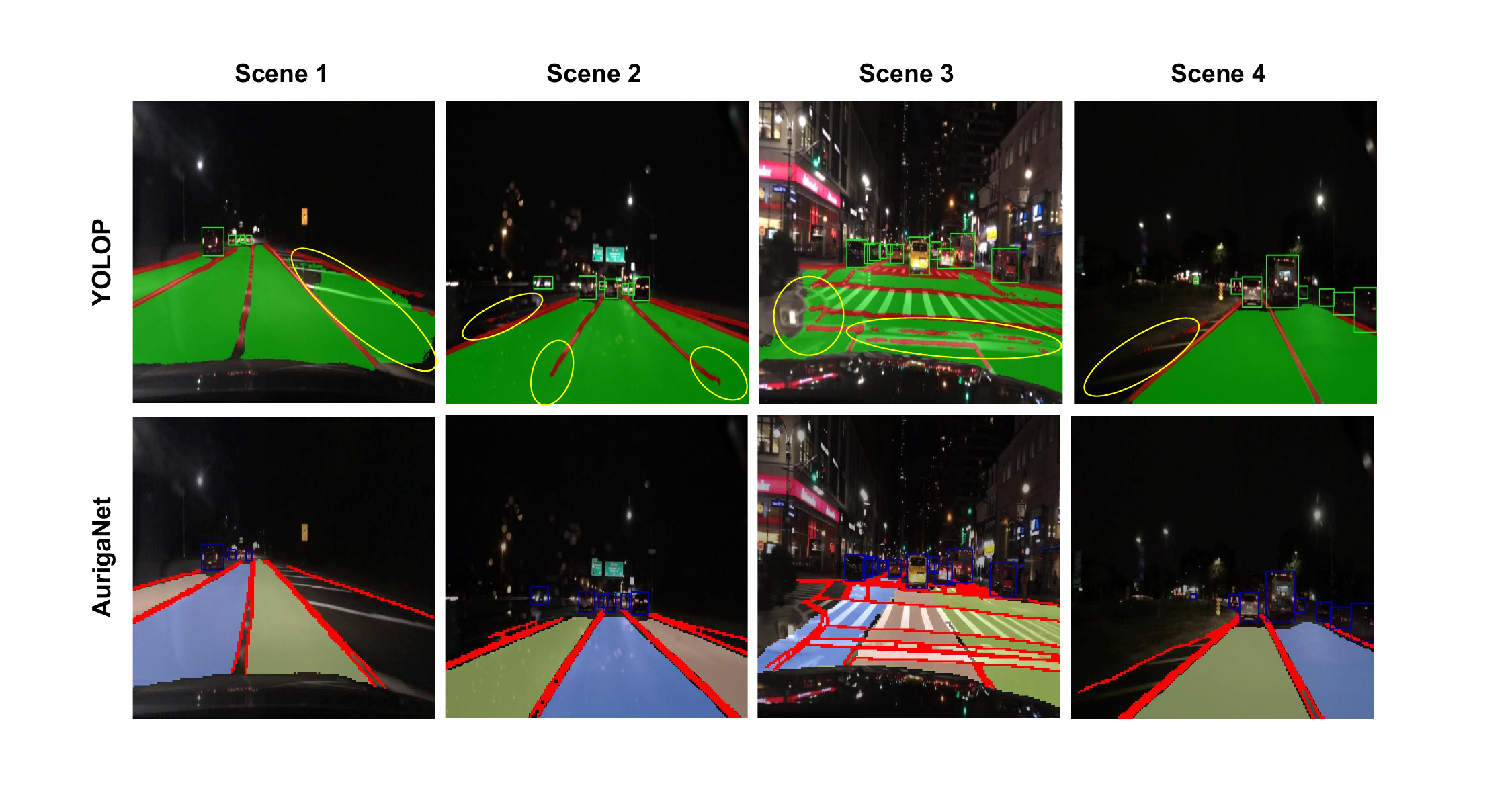}
\caption{The night-time results.}
\label{fig:9}
\end{figure}

\begin{figure}[H]
\centering
\includegraphics[width=0.85\textwidth]{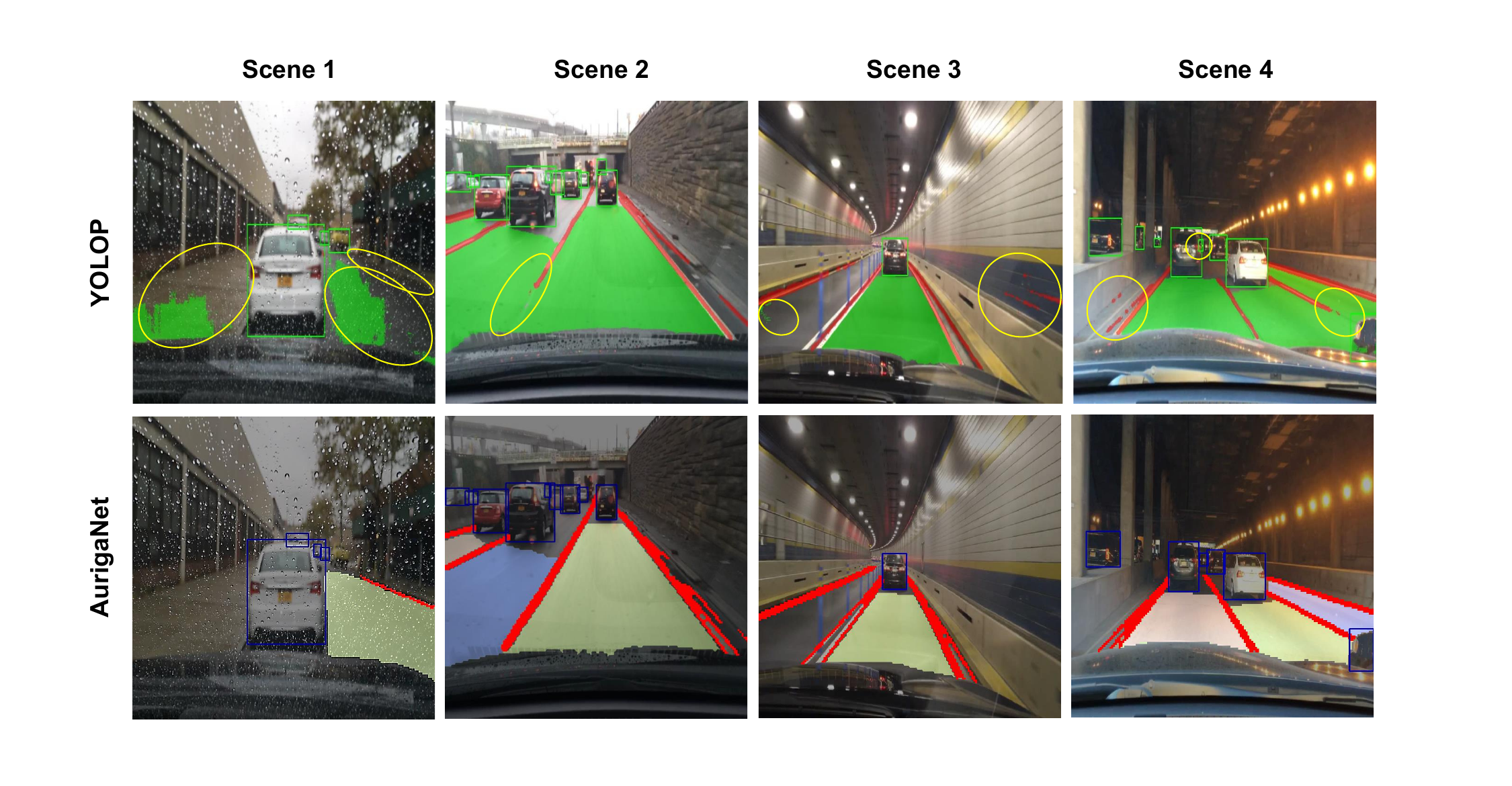}
\caption{The rain and tunnel results.}
\label{fig:10}
\end{figure}

We performed a qualitative comparison between the YOLOP and AurigaNet networks using randomly selected images from the BDD100K dataset. This comparison focuses on drivable area segmentation, lane detection, and traffic object detection tasks. Figures \ref{fig:8}, \ref{fig:9}, and \ref{fig:10} present the visualized results of the comparison. The yellow ovals highlight errors produced by the YOLOP model.
In Figure \ref{fig:8}, which depicts daytime scenes, YOLOP struggles with lane detection. Scenes 1 and 2 show inaccurate lane line predictions, while Scene 3 illustrates a false negative in which YOLOP mistakenly classifies the vehicle hood as a drivable area, in addition to failing at lane detection. Scene 4 further demonstrates YOLOP's poor performance in identifying the drivable area on the left side of the image and lane lines.
Figure \ref{fig:9} shows results in night-time environments. Scene 1 highlights a significant false negative, where YOLOP incorrectly labels the right side of the road as a drivable area, although it is a "Do Not Enter" area. Scenes 2, 3, and 4 depict YOLOP's consistent under-performance in lane detection.
In Figure \ref{fig:10}, which includes rainy and tunnel conditions, Scene 1 displays a false negative on the left side of the image, while Scene 2 exhibits poor lane detection. Scene 3 demonstrates YOLOP falsely identifying tunnel walls as lane lines and Scene 4 shows YOLOP inaccurately detecting part of a vehicle, along with poor lane detection.
Across all scenarios, AurigaNet consistently demonstrates superior accuracy in detecting drivable areas, lane lines, and objects, highlighting its ability to handle complex real-world driving environments while also accurately segmenting drivable area instances.

\section{Conclusion}
\label{conc}
In this paper, we introduced AurigaNet, an innovative real-time multi-task network architecture designed to address key challenges in autonomous driving perception. By leveraging the comprehensive BDD100K dataset, AurigaNet demonstrated superior performance in drivable path instance segmentation, lane detection, and object detection, outperforming state-of-the-art models across multiple metrics. Specifically, AurigaNet achieved 85.2\% IoU in drivable path segmentation, a 60.8\% IoU in lane detection (surpassing competing models by 30\%), and a 47.6\% mAP in traffic object detection. Table \ref{tabletotal} shows a comparison of AurigaNet with all state-of-the-art models.

The core innovation of AurigaNet lies in its ability to perform end-to-end instance segmentation with the mean-shift post-processing method, enhancing both accuracy and computational efficiency. Notably, drivable area instance segmentation is a crucial task for ADAS (Advanced Driver Assistance Systems) as it directly impacts path estimation and planning, which are essential for safe and efficient autonomous driving. Additionally, the integration of a discriminative loss function and deformable convolutions has further refined the network's capabilities, particularly in complex driving scenarios.

AurigaNet’s strong real-time performance on devices like the Jetson Orin NX demonstrates its practical applicability for autonomous driving systems, enhancing efficiency, safety, and scalability. Future refinements in architecture and optimization for various conditions will be key to unlocking its full potential, marking a significant step toward reliable and cost-effective autonomous driving solutions.

\newpage
\vspace{-40pt}
\vspace{-35pt}
\vspace{-32pt}

\FloatBarrier

\end{document}